\documentclass[runningheads]{llncs}

\usepackage{graphicx}
\usepackage{color}
\usepackage{amssymb}
\usepackage{amsmath}
\usepackage{bm}
\usepackage{url}
\newcommand{\bphi}{\boldsymbol{\phi}}
\newcommand{\btheta}{\bm{\theta}}
\newcommand{\bmu}{\bm{\mu}}
\newcommand{\bsigma}{\bm{\sigma}}
\newcommand{\bv}{\boldsymbol{v}}
\newcommand{\bu}{\boldsymbol{u}}
\DeclareMathOperator*{\argmax}{arg\,max}

\begin{document}
%
%
%
\title{Unsupervised Deep Learning for \\
Bayesian Brain MRI Segmentation}

\titlerunning{Unsupervised Deep Learning for Bayesian Brain MRI Segmentation}

\author{Adrian V. Dalca \inst{1,2} \and
Evan Yu\inst{3}\and
Polina Golland\inst{2}\and
Bruce Fischl\inst{1}\and \\
Mert R. Sabuncu\inst{3,4}\and
Juan Eugenio Iglesias\inst{1,2,5}}

\authorrunning{A. V. Dalca et al.}

\institute{Martinos Center for Biomedical Imaging, Massachusetts General Hospital, \\ Harvard Medical School \and
Computer Science and Artificial Intelligence Laboratory (CSAIL), \\Massachusetts Institute of Technology \and
Meinig School of Biomedical Engineering, Cornell University \and
School of Electrical and Computer Engineering, Cornell University \and
Centre for Medical Image Computing (CMIC), University College London}

\maketitle    
\begin{abstract}

Probabilistic atlas priors have been commonly used to derive adaptive and robust brain MRI segmentation algorithms. Widely-used neuroimage analysis pipelines rely heavily on these techniques, which are often computationally expensive. In contrast, there has been a recent surge of approaches that leverage deep learning to implement segmentation tools that are computationally  efficient at test time. However, most of these strategies rely on learning from manually annotated images. These supervised deep learning methods are therefore sensitive to the intensity profiles in the training dataset. To develop a deep learning-based segmentation model for a new image dataset (e.g., of different contrast), one usually needs to create a new labeled training dataset, which can be prohibitively expensive, or rely on suboptimal \textit{ad hoc} adaptation or augmentation approaches. In this paper, we propose an alternative  strategy that combines a conventional probabilistic atlas-based segmentation  with deep learning, enabling one to train a segmentation model for new MRI scans without the need for any manually segmented images. Our experiments include thousands of brain MRI scans and demonstrate that the proposed method achieves good accuracy for a brain MRI segmentation task for different MRI contrasts, requiring only approximately 15 seconds at test time on a GPU. The code is freely available at~\url{http://voxelmorph.mit.edu}.


\end{abstract}

%
%
\section{Introduction}

Bayesian segmentation of medical images, particularly in the context of brain MRI scans, is a well-studied problem. Most probabilistic models for image segmentation exploit atlas priors, and account for variations in contrast and imaging artifacts such as MR inhomogeneity~\cite{van1999automated,wells1996adaptive}. Most of the popular neuroimage processing pipelines rely on segmentation algorithms based on these ideas, including
FreeSurfer~\cite{fischl2002whole},
SPM~\cite{ashburner2005unified}, and FSL~\cite{patenaude2011bayesian}. While these tools achieve high robustness to changes in MRI contrast of the input scan, a significant drawback is that they are computationally demanding (e.g., 23 minutes using a multi-threaded setup, in a recent study~\cite{puonti2016fast}), which limits their deployment at scale and in time-sensitive applications. Therefore, there is a need for computationally efficient methods that are contrast-adaptive, requiring no  additional labeled training images to segment a new dataset.

Recently, there has been a surge in the application of deep learning (DL) techniques to medical image segmentation, often based on convolutional neural network (CNN) architectures that excel at learning contextually important multi-scale features. An advantage of these methods is their computational efficiency at test (segmentation) time, offering the potential to use automatic segmentation in new application areas, such as those involving very large test datasets~\cite{kamnitsas2017efficient,ronneberger2015}. Moreover, these algorithms can be combined with atlas priors for increased robustness~\cite{dalca2018anatomical,lee2019tetris,oktay2017anatomically}
However, DL based techniques are notoriously sensitive to changes in the image intensity data distribution.
For example, an upgrade to the MRI scanner or a change in the pulse sequence might alter contrast properties that can dramatically reduce the performance of a CNN-based segmentation model~\cite{jog2018pulse}. This issue can be alleviated via domain adaptation, which usually requires some amount of labeled training data for the new conditions, or data augmentation, which requires the user to simulate expected variations. However, even with additional data, these methods only partially close the gap with the fully supervised setting~\cite{pan2010survey}. Furthermore, the dependency on manually annotated datasets means that existing DL approaches are only applicable if enough resources are available to compile the required training data. This is often infeasible, for example in the context of continuously upgrading imaging technologies. 

In this paper, we consider the scenario in which we have a general probabilistic atlas prior and a collection of images with \textit{no manual delineations}. The probabilistic atlas is a volume where each voxel has an associated vector with the prior probabilities of observing the different segmentation labels at that location. Our approach assumes the availability of such an atlas (in brain imaging, they are readily available), and is independent of how it was created. For example, it could have been obtained by averaging a collection of manually annotated volumes of a different imaging modality. Alternatively, it could have been derived from an anatomical template, after applying spatial blurring to account for variability in location. 

Several recent methods tackle segmentation tasks in the presence of small training datasets. Most assume at least one manually segmented image from the same modality as the main task, and leverage data augmentations techniques and exploit priors to enable the use of supervised methods~\cite{chaitanya2019semi,zhao2019data}. Other methods require no labelled examples from the target modality, but leverage a large collection of segmentation maps from other datasets~\cite{dalca2018anatomical,joyce2018deep}.

The main contribution of this paper is the integration of mathematical ideas from the Bayesian segmentation literature with an unsupervised deep learning framework. 
Specifically, we assume a probabilistic model,
which requires estimation of scan-specific parameters comprising an atlas deformation and image intensity statistics. The estimation of the atlas warp has traditionally relied on classic deformable registration algorithms~\cite{sotiras2013deformable}, which are based on iterative, numerical optimization, and are therefore computationally expensive. Instead, we leverage recent advances in learning-based registration~\cite{balakrishnan2019tmi,dalca2018varreg,krebs2019learning,devos2017} to efficiently estimate the warp jointly with the intensity parameters.
We use a novel loss function, which is derived from the probabilistic model with Bayesian inference, and is thus principled and interpretable.
Integrating DL with Bayesian segmentation,  we attain two highly desirable features.
First, given a probabilistic atlas, the method is unsupervised, and hence contrast adaptive: given a new dataset with a previously unobserved MRI contrast (e.g., a change of pulse sequence in MRI acquisition), we train our network without the need to label any MRI scans. Second,  the segmentation is efficient and runs in approximately 15 seconds on a GPU.

%
%
\section{Method}

\subsection{Segmentation as Bayesian inference}
\label{sec:segmentAsBayes}

Let $\bm{I}$ represent the intensities of a 3D brain MRI scan, defined over a discrete domain $\Omega \subset \mathbb{R}^3$. 
Let $\bm{S}$ be a corresponding discrete segmentation into $L$ neuroanatomical labels. 
Bayesian segmentation relies on Bayes' rule to derive the posterior probability distribution of the segmentation~$\bm{S}$ given the input image~$\bm{I}$. Then, the  segmentation $\hat{\bm{S}}$ is estimated as the mode of this posterior:
\begin{equation}
\hat{\bm{S}} = \argmax_{\bm{S}} p(\bm{S} | \bm{I}) = \argmax_{\bm{S}} p(\bm{I} | \bm{S}) p(\bm{S}). \label{eq:Bayes} 
\end{equation}
The posterior distribution~$p(\bm{S} | \bm{I})$ depends on two terms: a prior~$p({\bm{S}})$ and a likelihood~$p(\bm{I} | \bm{S})$. This is in contrast to discriminative segmentation approaches, which model~$p(\bm{S} | \bm{I})$ directly. The prior represents knowledge about the spatial distribution of labels in the segmentation, and often has the form of a probabilistic atlas endowed with a deformation model. The likelihood models the relationship between the segmentation (i.e., underlying anatomy) and image intensities, including image artifacts such as noise and bias field. Both the prior and likelihood may have a set of associated parameters, which we define as $\btheta_S$ and $\btheta_I$, respectively. The former describes attributes such as label probabilities and atlas deformation, while the latter typically includes image intensity statistics as a function of label and possibly location. 

The likelihood parameters may be global for a training dataset, or estimated specifically for each test scan. Here we are interested in a subset of Bayesian segmentation models that follow the latter approach~\cite{ashburner2005unified,puonti2016fast,van1999automated,wells1996adaptive,zhang2001segmentation}, which enables these models to \emph{adapt} to the intensity characteristics of the input scans, making them robust to changes in MRI contrast.  Expanding Eq.~\eqref{eq:Bayes} to include model parameters, which we treat as random variables, yields:
\begin{equation}
\hat{\bm{S}} = \argmax_{\bm{S}} \int_{\btheta_S} \int_{\btheta_I} p(\bm{S} | \btheta_S, \btheta_I, \bm{I}) p( \btheta_S, \btheta_I | \bm{I}) d\bm{\theta_S} d\bm{\theta_I},
\end{equation}
which is intractable. A standard approximation is to use point estimates for the parameters. First, one estimates the mode of the posterior distribution for the parameters:
\begin{align}
\{ \hat{\btheta}_S , \hat{\btheta}_I \}  &= \argmax_{\{\btheta_S,\btheta_I\}} p(\btheta_S,\btheta_I | \bm{I})  \nonumber\\
&= 
 \argmax_{\{\btheta_S,\btheta_I\}}
 p(\btheta_S) p(\btheta_I ) \sum_{\bm{S}} p(\bm{I} | \bm{S}, \btheta_I) p(\bm{S} | \btheta_S), \label{eq:pointEstimates}
\end{align}
where we have assumed independence between the parameters of the prior and likelihood. The computation often requires estimating an atlas deformation in~$\btheta_S$ and intensity parameters in $\btheta_I$, and is typically achieved with a combination of numerical optimization and the Expectation Maximization (EM) algorithm~\cite{dempster1977maximum}. Given point estimates, the final segmentation is computed efficiently as:
\begin{equation}
\hat{\bm{S}} = \argmax_{\bm{S}} p(\bm{S} | \hat{\btheta}_S, \hat{\btheta}_I, \bm{I}) = \argmax_{\bm{S}}  p(\bm{I} | \bm{S}, \hat{\btheta}_I) p(\bm{S} | \hat{\btheta}_S), \label{eq:segmentation}
\end{equation}
and is often produced directly by the same EM algorithm.

\subsection{Proposed Model} 

Our model instantiation builds on existing work~\cite{ashburner2005unified,puonti2016fast,van1999automated}. The prior is defined by a given probabilistic atlas~$\bm{A}$, such that~$A (l, \bm{x})$  provides the probability of observing each neuroanatomical label $l=1,\ldots,L$ at each location $\bm{x}\in \Omega$. The atlas is deformed by a diffeomorphic transform~$\bphi$, parameterized by a stationary velocity field $\bv$, (i.e., $\bphi_v = \exp [\bv]$, see \cite{arsigny2006log}) that parametrizes the prior such that~$\btheta_S = \bv$.  Assuming independence over voxels:
\begin{equation}
p(\bm{S} | \btheta_S ; \bm{A}) = p (\bm{S} | \bv ; \bm{A}) = \prod_{j\in\Omega} A \Big( S_j , \bphi_v(\bm{x}_j) \Big),
\end{equation}
where $S_j$ is the segmentation at voxel~$j$, and $\bm{x}_j$ is its spatial location. 

We discourage strongly varying deformations by penalizing the spatial gradient~$\nabla \bu_v$ of displacement~$\bu_v$, where~$\bphi_v = Id + \bu_v$:
\begin{align}
p(\btheta_S ; \lambda) = p (\bv ; \lambda) \propto \exp[-\lambda \| \nabla \bu_v \|^2 ].
\end{align}
The hyperparameter $\lambda$ controls the weight for the atlas deformation penalty.

Conditioned on a segmentation, we assume that the observed intensities at different voxel locations are independent samples of Gaussian distributions:
\begin{equation}
p(\bm{I} | \bm{S},\btheta_I) = p(\bm{I} | \bm{S},\bmu,\bsigma^2)  = \prod_{j\in\Omega} \mathcal{N}(I_j ; \mu_{S_j} , \sigma^2_{S_j}),
\label{eq:lhood}
\end{equation}
where $\mathcal{N}(\cdot;\mu,\sigma^2)$ is the Gaussian distribution, $I_j$ is the image intensity at voxel $j$, and the likelihood parameters~$\btheta_I = \{\bmu, \bsigma^2\}$ are $L$ means~$\mu_l$ and variances~$\sigma_l^2$, each associated with a different label~$l$. We complete the model with a flat prior for these parameters: $p(\btheta_I)\propto 1$. The model can be easily extended to the multi-spectral case (i.e., inputs with multiple MRI contrasts) by replacing means and variances by mean vectors and covariance matrices, respectively.

\subsection{Learning}

\begin{figure}[t]
\centering
\includegraphics[width=1\textwidth]{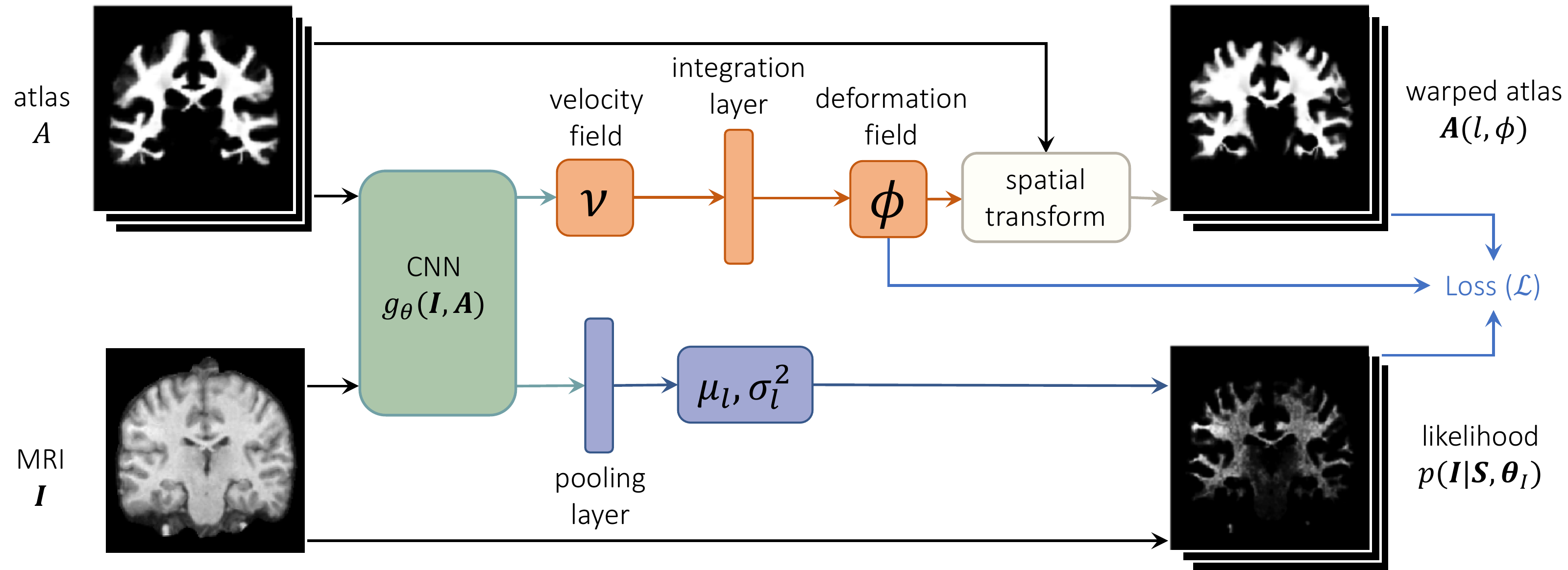}
\caption{\textbf{Method overview.} The network block~$g_\psi(\cdot,\cdot)$ outputs a stationary velocity field~$\bv$, enabling alignment of the probabilistic atlas to the input volume, and likelihood Gaussian parameters~$\bmu,\bsigma^2$, which yield likelihood maps for each label. 
} 
\label{fig:schematic}
\end{figure}

To avoid computationally expensive optimization typically required for 
maximum \textit{a posteriori} (MAP) estimation in Eq.~\eqref{eq:pointEstimates}, 
we propose to train a CNN to estimate these parameters directly from an input scan. Specifically, we design a CNN
$g_{\btheta_C} (\bm{I}, \bm{A}) = (\btheta_{S},\btheta_{I}) = (\bv,\bmu,\bsigma^2)$ with convolutional parameters~$\btheta_C$ that takes as input a scan~$\bm{I}$ and the probabilistic atlas $\bm{A}$, and outputs the model parameters~$\bv,\bmu,\bsigma^2$ for that scan. To learn the neural network parameters~$\bm{\theta}_c$, we use a pool of $N$ \textit{unlabeled} scans~$\{I^n\}_{n=1}^N$ to minimize the negative log posterior distribution of the image-specific parameters given the training images:
\begin{align}
- \sum_{n=1}^N &\log p(\bv^n,\bmu^{n},[\bsigma^2]^{n} | \bm{I}^n ; \bm{A}, \lambda) 
\label{eq:lossFunction}
\\
= &- \sum_{n=1}^N \sum_{j\in\Omega} \log \left[ \sum_{l=1}^L \mathcal{N}(I_{j}^{n} ; \mu_{l}^{n} , [\sigma^2_{l}]^{n})  A \Big( l , \bphi_{v_m}(\bm{x}_j)  \Big) \right] +  \lambda \| \nabla \bu_{v}^{n} \|^2  \nonumber \\
&- K(\lambda) + \text{const}, \nonumber %
\end{align}
where $K(\lambda)$ is a log-partition function that depends on the hyperparameter $\lambda$, and which does not affect the optimization.
We emphasize that the network outputs different parameters~$\bmu,\bsigma$, and~$\bv$ for each test image $\bm{I}$.

We design the neural network~$g_{\btheta_C}(\cdot,\cdot)$ based on a 3D UNet-style architecture~\cite{ronneberger2015} and the public VoxelMorph implementation~\cite{balakrishnan2019tmi}. The network consists of downsampling convolutional layers with 32 filters with 3x3 kernels, stride of 2, and  LeakyReLu activations, followed by mirror upsampling layers and skip-connections. From this point, an additional convolutional layer is used to output~$\bv$, a dense 3D velocity field defined over $\Omega$; and an additional pair of convolutional layers followed by a global max pooling operation to output the Gaussian parameters~$\bmu,\bsigma^2$. We compute~$\bphi = \exp(\bv)$ using a network integration layer that implements \textit{scaling and squaring}~\cite{arsigny2006log,dalca2018varreg,krebs2019learning}, enabling the computation of the loss regularization term. We warp the probabilistic atlas~$\bm{A}$ with a spatial transform layer. Combining the Gaussian parameters with the input image yields likelihood maps, which together with the warped atlas enable computation of the first term of the loss function (Fig.~\ref{fig:schematic}).

\subsection{Efficient segmentation}

Given a trained network and a new test subject, the network efficiently provides the image-specific parameter point estimates $\hat{\bv}$, and~$\hat{\btheta}_I$ via a single forward pass. The optimal segmentation can be efficiently computed for each voxel:
\begin{equation}
\hat{S}_j = \argmax_l \mathcal{N}(I_j ; \hat{\mu}_l , \hat{\sigma}^2_l) A \Big( l , \phi_{\hat{v}}(\bm{x}_j)  \Big).
\label{eq:segmentationVoxel}
\end{equation}
Both  terms in~Eq.~\eqref{eq:segmentationVoxel} are computed inside our GPU implementation (Fig.~\ref{fig:schematic}).

%
%
\section{Experiments and results}

\subsection{Data}

We evaluate our approach on three different image sets. The first dataset (``multi-site'') includes 8,332 T1-weighted scans from several public datasets: OASIS~\cite{marcus2007open}, ABIDE~\cite{di2014autism}, ADHD200~\cite{milham2012adhd}, MCIC~\cite{gollub2013mcic}, PPMI~\cite{marek2011parkinson}, HABS~\cite{dagley2015harvard}, and Harvard GSP~\cite{holmes2015brain}.
We randomly selected 7,332 scans to train and validate, and the remaining 1,000 were held out for testing. Manual delineations are not available for these scans, but we used automated segmentations produced by FreeSurfer~\cite{fischl2002whole} as a silver standard, for evaluation only.
The second dataset (``T1'') consist of 38  T1-weighted scans, used only for testing, each with 36 manually delineated brain structures~\cite{fischl2002whole}. 
The third dataset (``PD'') consists of eight proton density-weighted (PD) scans, manually segmented with the same protocol~\cite{fischl2004sequence}. 
All scans were preprocessed with FreeSurfer, 
including skull stripping, bias field correction, intensity normalization, affine registration to Talairach space, and resampling to 1 mm$^3$ isotropic resolution~\cite{fischl2012}.

\subsection{Experimental setup}
We perform three experiments, one for each dataset. 
In the first experiment, we fit our network to the 7,332 T1-weighted training scans of the multi-site dataset, and use the resulting model to segment the 1,000 test scans. Despite the lack of a manual gold standard, this experiment enables assessment of performance on a large, heterogeneous dataset. 
In a second experiment, we use the model already trained in the first experiment (i.e., on the 7,332 T1 scans) to segment scans from the separate T1 dataset. This experiment enables evaluation with manual ground truth on scans from a scanner and pulse sequence that were not observed by the neural network during training. 
In the third experiment, we train a network on the PD dataset, and then use it to segment those 8 PD scans. This is a different scenario than the first two experiments, since we learn to segment the test dataset directly. This experiment enables us to assess the ability of our algorithm to segment a substantially different MRI contrast, and to fit datasets of reduced size. In all experiments, we use our method with the publicly available atlas from~\cite{puonti2016fast}. We emphasize that all networks are trained in an unsupervised fashion, and segmentation maps are only used for evaluation.

\subsection{Baseline}

We compare our method to a reimplementation of~\cite{van1999automated}, which relies on an affine version of the aforementioned atlas and Gaussian likelihood functions. Specifically, the baseline method solves Eq.~\eqref{eq:lossFunction}, but with no deformation (i.e., $\bm{v}=\bm{u}=\bm{0}$, and  $\phi_v = Id$), and the model parameters are estimated with the EM algorithm. Since the model does not include deformation, using the nonrigid version of the atlas would yield very low performance.

\begin{figure}[t]
\centering
\includegraphics[width=0.49\textwidth]{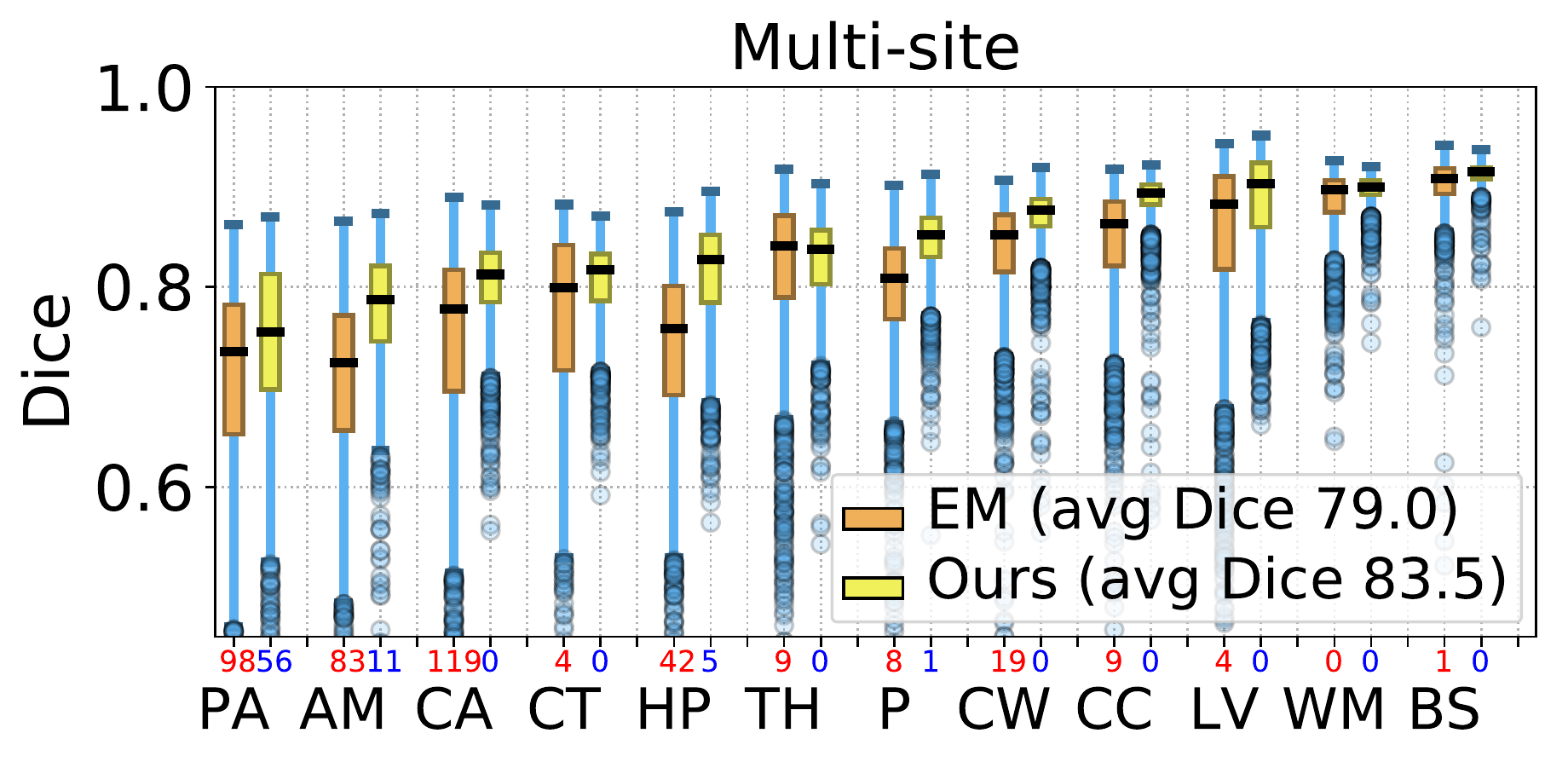}
\includegraphics[width=0.49\textwidth]{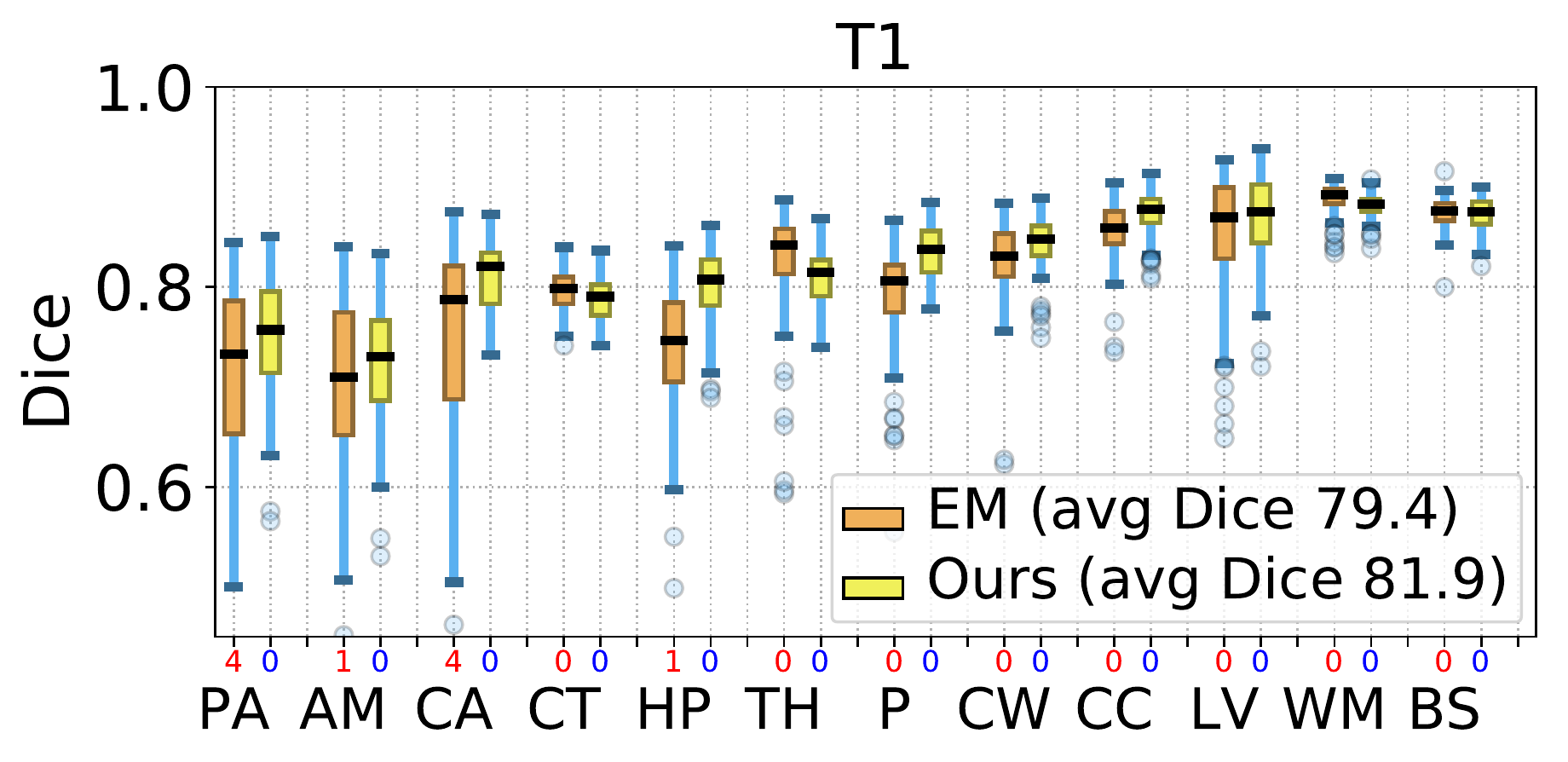}\newline
\includegraphics[width=0.49\textwidth]{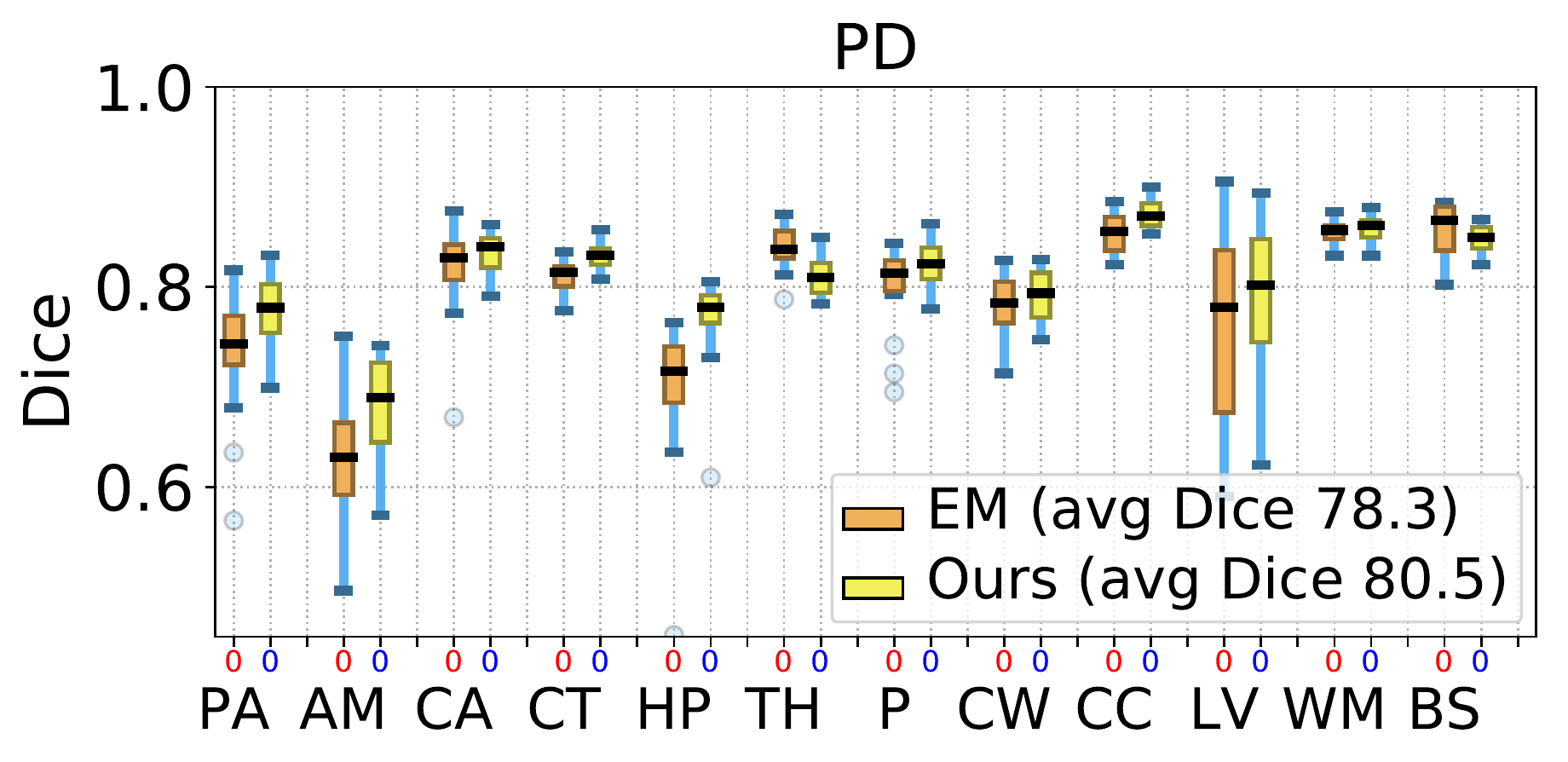}
\includegraphics[width=0.49\textwidth]{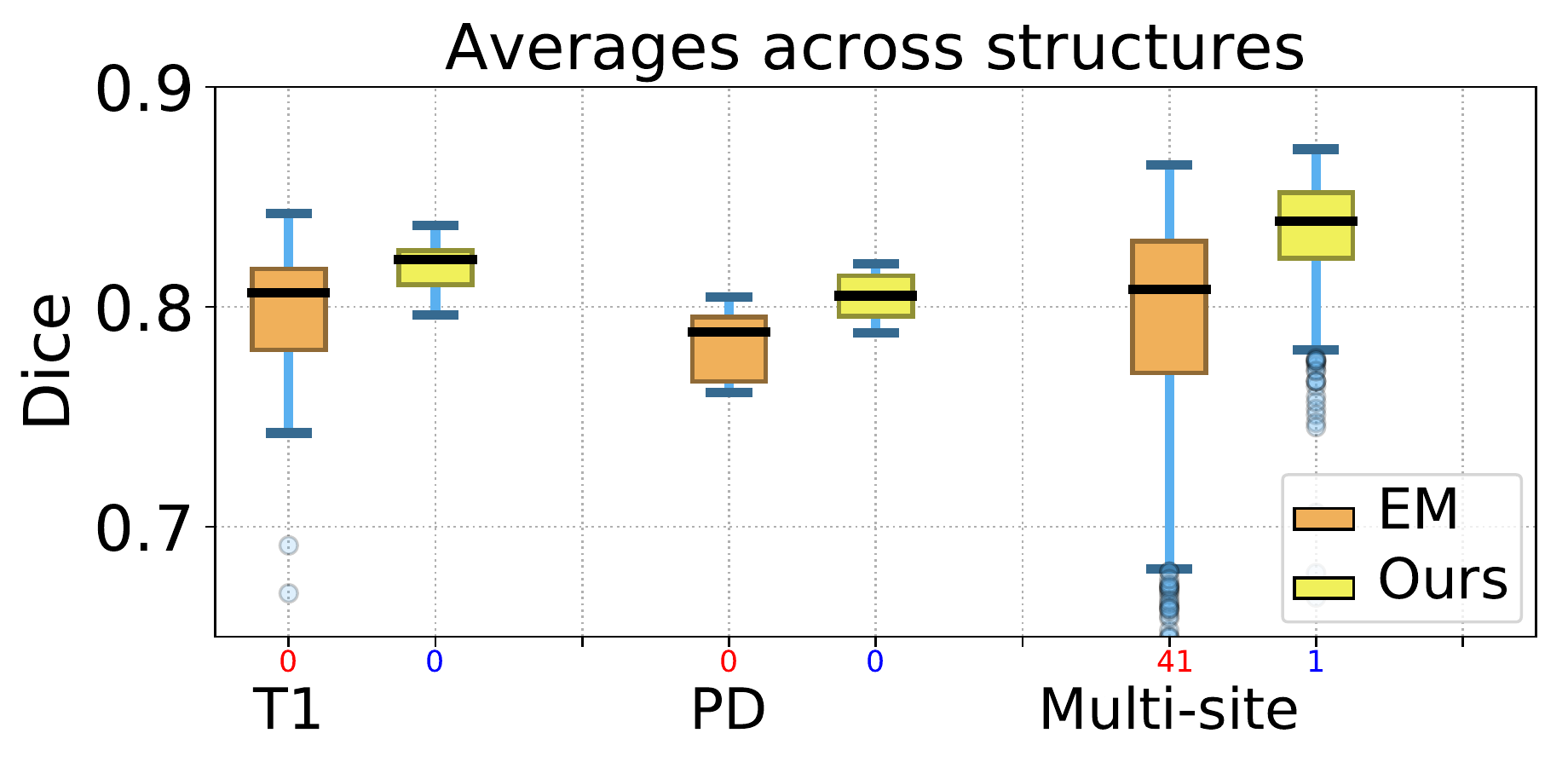}
\caption{\textbf{Segmentation Statistics.} 
Dice scores for: cerebral cortex (CT) and white matter (WM); lateral ventricle (LV); cerebellar cortex (CC) and white matter (CW); thalamus (TH); caudate (CA); putamen (P); pallidum (PA); brainstem (BS); hippocampus (HP); and amygdala (AM). Scores of contralateral structures are averaged. The number of outliers  under the $x$ axis is shown in red (baseline) and blue (ours).
} 
\label{fig:results}
\end{figure}

\subsection{Evaluation} 
We used Dice scores for a subset of structures of interest (Fig.~\ref{fig:results}). We quantify the results on these structures, and also focus on deep structures such as the hippocampus, which is the target of many neuroimaging studies due to its significance in dementia.

\subsection{Implementation Details}
We group anatomical labels with similar intensity properties into eleven merged labels to force groups of original labels to share Gaussian parameters, increasing robustness~\cite{puonti2016fast}. Specifically, we group: contralateral structures (in general), gray matter structures (cerebral gray matter, hippocampus, amygdala, caudate, accumbens), and cerebrospinal fluid structures. 

We implement our method using Keras~\cite{chollet2015} with a Tensorflow~\cite{abadi2016tensorflow} backend and the ADAM optimizer~\cite{kingma2014}. We predict the velocity field $\bv$ and resulting deformation field~$\bphi$ at every second voxel in each dimension, due to memory constraints. We linearly interpolate to obtain a final dense deformation field. To set $\lambda$, the only free parameter of our framework, we visually evaluated segmentation results for several validation subjects (held out from the training dataset), and set~$\lambda=10$ in all experiments.

\begin{figure}[t]
\centering
\includegraphics[width=0.99\textwidth]{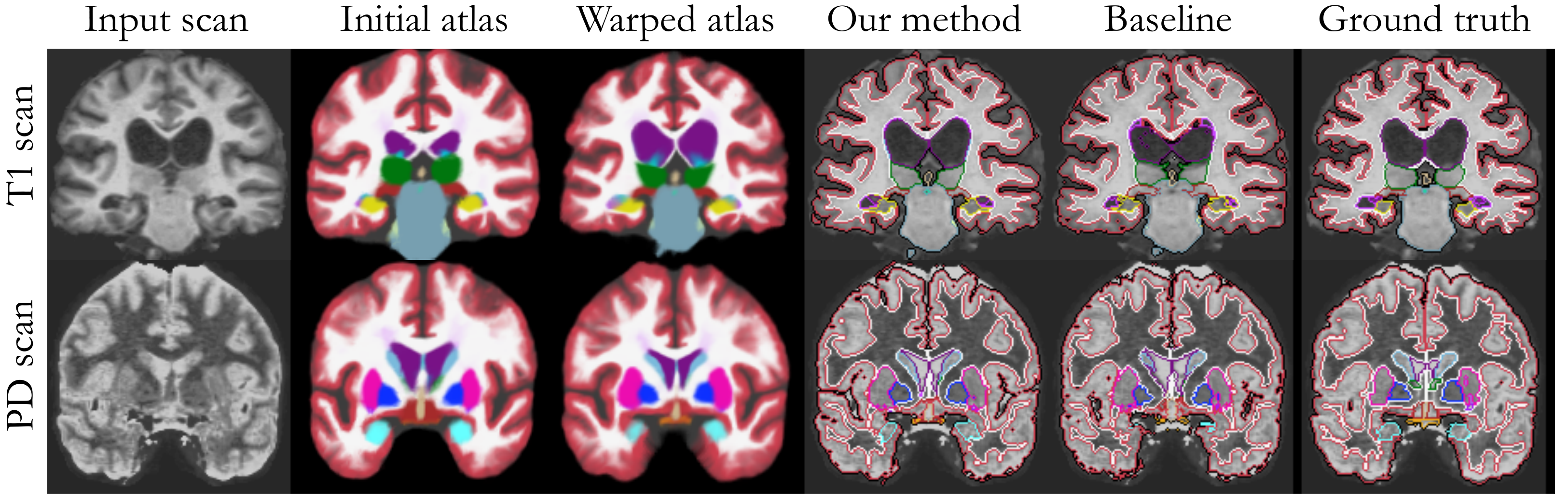}
\caption{\textbf{Example Results.} Coronal slices of two scans (one from each of the T1 and PD datasets), along with the initial and deformed probabilistic atlas, and corresponding segmentations. In the atlas, the color of each pixel is a combination of the colors of different labels, weighted by their probabilities. In the segmentations, we show the contour of the labels in the corresponding colors. We use the FreeSurfer color map~\cite{fischl2012}. 
} 
\label{fig:resultExamples}
\end{figure}

\subsection{Results}

Our method requires only 15 seconds per scan on an NVIDIA Titan Xp GPU. Fig.~\ref{fig:results} reports segmentation statistics for all experiments. Our method achieves considerably higher Dice scores than the baseline on the multi-site dataset (average over all structures 83.5\% \textit{vs.} 79.0\%), particularly in deep brain structures, such as the hippocampi (81.1\% \textit{vs.} 73.1\%). Moreover, it largely reduces the number of outliers with very poor segmentation (e.g., there are over 100 cases with Dice lower than 50\% in the caudate for the baseline approach, and none for our method).
In the T1 dataset, the test intensity distribution is slightly different that of the training dataset.
However, our approach successfully generalizes and outperforms the baseline (average 81.9\% \textit{vs.} 79.4\%, hippocampi 79.9\% \textit{vs.} 73.5\%). 
The results of the third experiment illustrate the ability of our method to adapt to contrasts other than T1, even when the data are limited, and outperform the baseline (average 80.5\% \textit{vs.} 78.3\%, hippocampi 76.6\% \textit{vs.} 69.8\%). 
 
Figure~\ref{fig:resultExamples} shows two segmentations from the T1 and PD datasets. In the T1 scan, the atlas successfully deforms to match the large ventricles of the  subject, producing more accurate segmentations than the baseline -- not only for the ventricles (purple), but also for surrounding structures, e.g., the thalami (green). In the PD scan, despite the small dataset, our method manages to segment all structures  including the amygdalae (light blue), which are missed by the baseline.

%
%
\section{Conclusion}
We propose a principled approach for unsupervised segmentation, which enables training a CNN for a dataset without the need for \textit{any} manually annotated images. The likelihood model may be extended to incorporate more complex functions (such as mixtures of Gaussians) and artifacts such as partial voluming and bias field. In addition to segmentations, the method produces a dense nonlinear deformation field that is a useful output by itself, e.g., for tensor-based morphometry. Using a large dataset, we demonstrate that the proposed approach achieves state-of-the-art accuracy for \textit{unsupervised} brain MRI segmentation in different MRI contrasts. Our method runs in under 15 seconds on a GPU, facilitating deployment on large studies and in time-sensitive applications.


\section*{Acknowledgement}
This work was supported by the European Research Council (Starting Grant 677697, project ``BUNGEE-TOOLS'', awarded to JEI); 
the BRAIN Initiative Cell Census Network (U01-MH117023); 
NIH (R21-AG050122, P41-EB015902, 5U01-MH093765, R01LM012719,
R01AG053949, and 1R21AG050122);  
NIBIB (P41-EB015896, 1R01-EB023281, R01-EB006758, R21-EB018907, R01-EB019956), 
NIA (5R01-AG008122, R01-AG016495), NIDDK (1R21-DK-108277-01), 
NINDS (R01-NS0525851, R21-NS072652, R01-NS070963, R01-NS083534, 5U01-NS086625, 5U24-NS10059103); 
NSF NeuroNex Grant grant 1707312, CAREER 1748377;
Wistron Corp.; SIP; and AWS; 
and was made possible by the resources provided by Shared Instrumentation Grants 1S10RR023401, 1S10RR019307, and 1S10RR023043. In addition, BF has a financial interest in CorticoMetrics, a company whose medical pursuits focus on brain imaging and measurement technologies. BF's interests were reviewed and are managed by Massachusetts General Hospital and Partners HealthCare in accordance with their conflict of interest policies.

\bibliographystyle{splncs04}
\bibliography{references}

\begin{thebibliography}{10}
\providecommand{\url}[1]{\texttt{#1}}
\providecommand{\urlprefix}{URL }
\providecommand{\doi}[1]{https://doi.org/#1}

\bibitem{abadi2016tensorflow}
Abadi, M., Barham, P., Chen, J., Chen, Z., Davis, A., Dean, J., Devin, M.,
  Ghemawat, S., Irving, G., Isard, M., et~al.: Tensorflow: A system for
  large-scale machine learning. In: 12th $\{$USENIX$\}$ Symposium on Operating
  Systems Design and Implementation ($\{$OSDI$\}$ 16). pp. 265--283 (2016)

\bibitem{arsigny2006log}
Arsigny, V., Commowick, O., Pennec, X., Ayache, N.: A log-euclidean framework
  for statistics on diffeomorphisms. In: MICCAI. pp. 924--931. Springer (2006)

\bibitem{ashburner2005unified}
Ashburner, J., Friston, K.: Unified segmentation. Neuroimage  \textbf{26},
  839--851 (2005)

\bibitem{balakrishnan2019tmi}
Balakrishnan, G., Zhao, A., Sabuncu, M., Guttag, J., Dalca, A.V.: Voxelmorph: A
  learning framework for deformable medical image registration. IEEE:TMI
  (2019)

\bibitem{chaitanya2019semi}
Chaitanya, K., Karani, N., Baumgartner, C., Konukoglu, E.: Semi-supervised and
  task-driven data augmentation. arXiv preprint arXiv:1902.05396  (2019)

\bibitem{chollet2015}
Chollet, F.: Keras. \url{https://github.com/fchollet/keras} (2015)

\bibitem{dagley2015harvard}
Dagley, A., LaPoint, M., Huijbers, W., Hedden, T., McLaren, D.G., Chatwal,
  J.P., Papp, K.V., Amariglio, R.E., Blacker, D., Rentz, D.M., et~al.: Harvard
  aging brain study: dataset and accessibility. NeuroImage  (2015)

\bibitem{dalca2018varreg}
Dalca, A.V., Balakrishnan, G., Guttag, J., Sabuncu, M.: Unsupervised learning
  for fast probabilistic diffeomorphic registration. MICCAI  \textbf{11070},
  729--738. (2018)

\bibitem{dalca2018anatomical}
Dalca, A.V., Guttag, J., Sabuncu, M.R.: Anatomical priors in convolutional
  networks for unsupervised biomedical segmentation. In: Proceedings of the
  IEEE Conference on Computer Vision and Pattern Recognition. pp. 9290--9299
  (2018)

\bibitem{dempster1977maximum}
Dempster, A.P., Laird, N.M., Rubin, D.B.: Maximum likelihood from incomplete
  data via the {EM} algorithm. J. Royal Statistical Society: Series B
  \textbf{39}(1),  1--22 (1977)

\bibitem{di2014autism}
Di~Martino, A., et~al.: The autism brain imaging data exchange: towards a
  large-scale evaluation of the intrinsic brain architecture in autism.
  Molecular psychiatry  \textbf{19}(6),  659--667 (2014)

\bibitem{fischl2012}
Fischl, B.: Freesurfer. Neuroimage  \textbf{62}(2),  774--781 (2012)

\bibitem{fischl2002whole}
Fischl, B., Salat, D.H., Busa, E., Albert, M., Dieterich, M., et~al.: Whole
  brain segmentation: automated labeling of neuroanatomical structures in the
  human brain. Neuron  \textbf{33}(3),  341--355 (2002)

\bibitem{fischl2004sequence}
Fischl, B., Salat, D.H., Van Der~Kouwe, A.J., Makris, N., S{\'e}gonne, F.,
  Quinn, B.T., Dale, A.M.: Sequence-independent segmentation of magnetic
  resonance images. Neuroimage  \textbf{23},  S69--S84 (2004)

\bibitem{gollub2013mcic}
Gollub, R.L., Shoemaker, J.M., King, M.D., White, T., Ehrlich, S., Sponheim,
  S.R., Clark, V.P., Turner, J.A., Mueller, B.A., Magnotta, V., et~al.: The
  {MCIC} collection: a shared repository of multi-modal, multi-site brain image
  data from a clinical investigation of schizophrenia. Neuroinformatics
  \textbf{11}(3),  367--388 (2013)

\bibitem{holmes2015brain}
Holmes, A.J., Hollinshead, M.O., O’Keefe, T.M., Petrov, V.I., Fariello, G.R.,
  Wald, L.L., Fischl, B., Rosen, B.R., Mair, R.W., Roffman, J.L., et~al.: Brain
  genomics superstruct project initial data release with structural,
  functional, and behavioral measures. Scientific data  \textbf{2} (2015)

\bibitem{jog2018pulse}
Jog, A., Fischl, B.: Pulse sequence resilient fast brain segmentation. In:
  MICCAI. pp. 654--662. Springer (2018)

\bibitem{joyce2018deep}
Joyce, T., Chartsias, A., Tsaftaris, S.A.: Deep multi-class segmentation
  without ground-truth labels. MIDL  (2018)

\bibitem{kamnitsas2017efficient}
Kamnitsas, K., Ledig, C., Newcombe, V., Simpson, J., Kane, A.D., Menon, D.,
  Rueckert, D., Glocker, B.: Efficient multi-scale {3D} {CNN} with fully
  connected {CRF} for accurate brain lesion segmentation. Medical image
  analysis  \textbf{36},  61--78 (2017)

\bibitem{kingma2014}
Kingma, D.P., Ba, J.: {ADAM}: A method for stochastic optimization. arXiv
  preprint arXiv:1412.6980  (2014)

\bibitem{krebs2019learning}
Krebs, J., e~Delingette, H., Mailh{\'e}, B., Ayache, N., Mansi, T.: Learning a
  probabilistic model for diffeomorphic registration. IEEE transactions on
  medical imaging  (2019)

\bibitem{lee2019tetris}
Lee, M.C.H., Petersen, K., Pawlowski, N., Glocker, B., Schaap, M.: {TETRIS}:
  Template transformer networks for image segmentation with shape priors,
  {IEEE} Trans. Med. Imaging, accepted

\bibitem{marcus2007open}
Marcus, D.S., Wang, T.H., Parker, J., Csernansky, J.G., Morris, J.C., Buckner,
  R.L.: Open access series of imaging studies (oasis): cross-sectional mri data
  in young, middle aged, nondemented, and demented older adults. Journal of
  cognitive neuroscience  \textbf{19}(9),  1498--1507 (2007)

\bibitem{marek2011parkinson}
Marek, K., Jennings, D., Lasch, S., Siderowf, A., Tanner, C., Simuni, T.,
  Coffey, C., Kieburtz, K., Flagg, E., Chowdhury, S., et~al.: The parkinson
  progression marker initiative (ppmi). Progress in neurobiology
  \textbf{95}(4),  629--635 (2011)

\bibitem{milham2012adhd}
Milham, M.P., Fair, D., Mennes, M., Mostofsky, S.H., et~al.: The {ADHD}-200
  consortium: a model to advance the translational potential of neuroimaging in
  clinical neuroscience. Frontiers in systems neuroscience  \textbf{6}, ~62
  (2012)

\bibitem{oktay2017anatomically}
Oktay, O., Ferrante, E., Kamnitsas, K., Heinrich, M., Bai, W., Caballero, J.,
  Cook, S.A., De~Marvao, A., Dawes, T., O‘Regan, D.P., et~al.: Anatomically
  constrained neural networks {(ACNNs)}: application to cardiac image
  enhancement and segmentation. IEEE Trans. Med. Imaging  \textbf{37}(2),
  384--395 (2017)

\bibitem{pan2010survey}
Pan, S.J., Yang, Q.: A survey on transfer learning. IEEE:TKDE  \textbf{22},
  1345--59 (2010)

\bibitem{patenaude2011bayesian}
Patenaude, B., Smith, S., Kennedy, D., Jenkinson, M.: A bayesian model of shape
  and appearance for subcortical brain segmentation. Neuroimage  \textbf{56},
  907--922 (2011)

\bibitem{puonti2016fast}
Puonti, O., Iglesias, J., Van~Leemput, K.: Fast sequence-adaptive whole-brain
  segmentation using parametric bayesian modeling. NeuroImage  \textbf{143},
  235--249 (2016)

\bibitem{ronneberger2015}
Ronneberger, O., Fischer, P., Brox, T.: U-net: Convolutional networks for
  biomedical image segmentation. In: MICCAI. pp. 234--241. Springer (2015)

\bibitem{sotiras2013deformable}
Sotiras, A., Davatzikos, C., Paragios, N.: Deformable medical image
  registration: A survey. IEEE:TMI  \textbf{32}(7), ~1153 (2013)

\bibitem{van1999automated}
Van~Leemput, K., Maes, F., Vandermeulen, D., Suetens, P.: Automated model-based
  tissue classification of {MR} images of the brain. IEEE:TMI  \textbf{18},
  897--908 (1999)

\bibitem{devos2017}
de~Vos, B., Berendsen, F., Viergever, M., Staring, M., I{\v{s}}gum, I.:
  End-to-end unsupervised deformable image registration with a {CNN}. DLMIA,
  pp. 204--212 (2017)

\bibitem{wells1996adaptive}
Wells, W.M., Grimson, W.E.L., Kikinis, R., Jolesz, F.A.: Adaptive segmentation
  of {MRI} data. IEEE:TMI  \textbf{15}(4),  429--442 (1996)

\bibitem{zhang2001segmentation}
Zhang, Y., Brady, M., Smith, S.: Segmentation of brain {MR} images through a
  hidden markov random field model and the expectation-maximization algorithm.
  IEEE:TMI  \textbf{20}(1),  45--57 (2001)

\bibitem{zhao2019data}
Zhao, A., Balakrishnan, G., Durand, F., Guttag, J.V., Dalca, A.V.: Data
  augmentation using learned transformations for one-shot medical image
  segmentation. In: Proceedings of the IEEE Conference on Computer Vision and
  Pattern Recognition. pp. 8543--8553 (2019)

\end{thebibliography}

\end{document}